\documentclass[runningheads, envcountsame, a4paper]{llncs} 
\usepackage[ruled,linesnumbered]{algorithm2e}
\usepackage{mathtools, amssymb}
\usepackage{booktabs}
\usepackage{multirow, multicol}
\usepackage{placeins}
\usepackage[pdftex]{graphicx} 
\usepackage{epstopdf} 
\usepackage{cite}
\usepackage[hyphens]{url}
\usepackage{bibnames}
\usepackage[misc]{ifsym}

\begin{document}
\title{Non-Exhaustive Learning Using Gaussian Mixture Generative Adversarial Networks
}
\titlerunning{Non-Exhaustive Gaussian Mixture Generative Adversarial Networks}
\author{Jun Zhuang {\Letter} \orcidID{0000-0002-7142-2193} \\
Mohammad Al Hasan {\Letter} \orcidID{0000-0002-8279-1023}}
\authorrunning{Jun Zhuang \and Mohammad Al Hasan}
\institute{Indiana University-Purdue University Indianapolis, Indianapolis, IN, 46202, USA
\email{junz@iu.edu, alhasan@iupui.edu}
}
\toctitle{Non-Exhaustive Learning Using Gaussian Mixture Generative Adversarial Networks}
\tocauthor{Jun~Zhuang, Mohammad~Al~Hasan}
\maketitle
\begin{abstract}
Supervised learning, while deployed in real-life scenarios, often encounters instances of unknown classes. Conventional algorithms for training a supervised learning model do not provide an option to detect such instances, so they miss-classify such instances with 100\% probability.
Open Set Recognition (OSR) and Non-Exhaustive Learning (NEL) are potential solutions to overcome this problem. Most existing methods of OSR first classify members of existing classes and then identify instances of new classes. However, many of the existing methods of OSR only makes a binary decision, i.e., they only identify the existence of the unknown class. Hence, such methods cannot distinguish test instances belonging to incremental unseen classes. On the other hand, the majority of NEL methods often make a parametric assumption over the data distribution, which either fail to return good results, due to the reason that real-life complex datasets may not follow a well-known data distribution.
In this paper, we propose a new online non-exhaustive learning model, namely, Non-Exhaustive Gaussian Mixture Generative Adversarial Networks (NE-GM-GAN) to address these issues. Our proposed model synthesizes Gaussian mixture based latent representation over a deep generative model, such as GAN, for incremental detection of instances of emerging classes in the test data. Extensive experimental results on several benchmark datasets show that NE-GM-GAN significantly outperforms the state-of-the-art methods in detecting instances of novel classes in streaming data.

\end{abstract}

\section{Introduction}
\label{sec:intro}
Numerous machine learning models are supervised, relying substantially on labeled datasets. In such datasets, the labels of training instances enable a supervised model to learn the correlation between the labels and the patterns in the features, thus helping the model to achieve the desired performance in different kinds of classification or recognition tasks. 
However, many realistic machine learning problems originate in non-stationary environments where instances of unseen classes may emerge naturally. The presence of such instances weakens the robustness of conventional machine learning algorithms, as these algorithms do not account for the instances from unknown classes, either in the train or the test environments.
To overcome this challenge, a series of related research activities has become popular in recent years; examples include anomaly detection (AD) \cite{liu2008isolation, manevitz2001one, schlegl2017unsupervised, zong2018deep}, few-shot learning (FSL) \cite{koch2015siamese, ravi2016optimization}, zero-shot learning (ZSL) \cite{socher2013zero, palatucci2009zero}, open set recognition (OSR) and open-world classification (OWC) \cite{scheirer2014probability, mensink2013distance, bendale2015towards, bendale2016towards, ge2017generative, neal2018open, jo2018open, yang2019open,  liu2019large, wang2018iterative, geng2020collective, hassen2020learning, oza2019c2ae, perera2020generative}. Collectively, each of these works belongs to one of the four different categories~\cite{geng2018recent}, differing on the kind of instances observed by the model during train and test.
If $L$ refers to labeling and $I$ refers to self-information (e.g., semantic information in image dataset), the categories $C$ can be denoted as the Cartesian product of two sets $L$ and $I$, as shown below:
\begin{equation}
  C = L \times I = \{(l, i)~:~l \in L \ \& \ i \in I \},
\end{equation}
both $L$ and $I$ have two elements: known (K) and unknown (U). Thus, there are four categories in $C$: (K, K), (K, U), (U, K), (U, U). For example, (U, U) refers to the learning problem in which instances belonging to unknown classes having no self-information.

Conventional supervised learning task belongs to the first category, as for such a task all instances in train and test datasets belong to (K, K).
The anomaly detection (AD) task, a.k.a. one-class classification or outlier detection, detects a few (U, U) instances from the majority of (K, K) instances; for AD, the (U, U) instances may only (but not necessary) exist in the test set. 
FSL and ZSL are employed to identify (U, K) instances in the test set. The main difference between FSL and ZSL is that the training set of FSL contains a limited number of (U, K) instances while for the case of ZSL, the number of (U, K) instances in the train set is zero. In other words, ZSL identifies (U, K) instances in the test set only by associating (K, K) instances with (U, K) instances through self-information. 
Finally, works belonging to open set recognition (OSR) identify (U, U) instances in the test set. These works are the most challenging; unlike AD, whose objective is to detect only one class (outlier), OSR handles both (K, K) and (U, U) in the test set. 
Similar to OSR, OWC also incrementally learns the new classes and rejects the unseen class. 
Nevertheless, most existing methods of OSR or OWC do not distinguish the test instances among incremental unseen classes, which is more close to the realistic scenario. The scope of our work falls in the OSR category which only deals with (K, K) and (U, U) instances. 
In Table~\ref{tab:background}, we present a summary of the discussion of this paragraph.

\begin{table*}
\centering
  \caption{The Background of Related Tasks (Conv. for Conventional Method)}
  \label{tab:background}
  \begin{tabular}{cccc} 
    \toprule
    Tasks & Training Set & Testing Set & GOAL \\
    \midrule
    Conv. & (K, K) & (K, K) & Supervised learning with (K, K) \\
    \hline
    AD & (K, K) w./wo. outliers & (K, K) w. outliers & Detect outliers \\
    \hline
    FSL & (K, K) w. limited (U, K) & (U, K) & Identify (U, K) in test set \\
    \hline
    ZSL & (K, K) w. self-info. & (U, K) & Identify (U, K) in test set \\
    \hline
    OSR & (K, K) & (K, K) \& (U, U) & Distinguish (U, U) from (K, K) \\
    \hline
    NEL & (K, K) & (K, K) \& (U, U) & Incrementally learn (U, U) \\
  \bottomrule
\end{tabular}
\end{table*}

Some works belonging to OSR have also been referred as Non-Exhaustive Learning (NEL). The term, Non-Exhaustive, means that the training data does not have instances of all classes that may be expected in the test data. The majority of early research works of NEL employ Bayesian methods with Gaussian mixture model (GMM) or infinite Gaussian mixture model (IGMM) \cite{rasmussen2000infinite, zhang2016bayesian}. However, these works suffer from some limitations; for instance, they assume that the data distribution in each class follows a mixture of Gaussian, which may not be true in many realistic datasets. Also, in the case of GMM, its ability to recognize unknown classes depends on the number of initial clusters that it uses. IGMM can mitigate this restriction by allowing cluster count to grow on the fly, but the inference mechanism of IGMM is time-consuming, no matter what kind of sampling method it uses for inferring the probabilities of the posterior distribution.

To address these issues, in this work we propose a new non-exhaustive learning model, Non-exhaustive Gaussian mixture Generative Adversarial Networks (NE-GM-GAN), which synthesizes the Bayesian method and deep learning technique. Comparing to the existing methods for OSR, our proposed method has several advantages: First, NE-GM-GAN takes multi-modal prior as input to better fit the real data distribution; Second, NE-GM-GAN can deal with class-imbalance problem with end-to-end offline training; Finally, NE-GM-GAN can achieve accurate and robust online detection on large sparse dataset while avoiding noisy distraction. Extensive experiments demonstrate that our proposed model has superior performance over competing methods on benchmark datasets.
The contribution of this paper can be summarized as follows:

$\bullet$\  We propose a new model for non-exhaustive learning, namely NE-GM-GAN, which can detect novel classes in online test data accurately and defy the class-imbalance problem effectively.

$\bullet$\  NE-GM-GAN integrates Bayesian inference with the distance-based and the threshold-based method to estimate the number of emerging classes in the test data. It also devises a novel scoring method to distinguish the UCs (unknown classes) from KCs (known classes).

$\bullet$\ Extensive experiments on four datasets (3 real and 1 synthetic) demonstrate that our model is superior to existing methods for accurate and robust online detection of emerging classes in streaming data.

\section{Related Work}
\label{sec:rewk}
Anomaly detection (AD) basically can be divided into two categories, conventional methods, and deep learning techniques. Majority of conventional methods widely focus on distance-based approaches \cite{scholkopf2000support, manevitz2001one}, reconstruction-based approaches \cite{hubert2005robpca}, and unsupervised clustering. 
Deep learning techniques usually include autoencoder and GAN. An autoencoder identifies the outlier instances through reconstruction loss \cite{zong2018deep}. GAN has also been used as another means for computing reconstruction loss and then identifying anomalies \cite{schlegl2017unsupervised, zenati2018adversarially}. In our approach, we use bi-directional GAN (BiGAN) with multi-modal prior distribution to improve the performance of UCs extraction.

AD mainly detects one class of anomalies whereas realistic data often contains multiple UCs. OSR is the right technique that solves this kind of problem. According to \cite{geng2018recent}, OSR models are categorized into two types, discriminative and generative. The first type includes SVM-based methods~\cite{scheirer2014probability} and distance-based method~\cite{bendale2015towards, bendale2016towards}. A collection of recent OSR works venture towards the generative direction \cite{ge2017generative, neal2018open, jo2018open, yang2019open}. A subset of OSR methods, named NEL, mainly employ Bayesian methods, such as infinite Gaussian mixture model (IGMM) \cite{rasmussen2000infinite} to learn the UCs. For example, Zhang et al. \cite{zhang2016bayesian} use a non-parametric Bayesian framework with different posterior sampling strategies, such as one sweep Gibbs sampling, for detecting novel classes in online name disambiguation. However, IGMM-type methods can only handle small datasets that follow Gaussian distribution. To address this issue, we propose a novel algorithm that can achieve high accuracy on the large sparse dataset, which does not necessarily follow Gaussian distribution.

\section{Background}
\label{sec:bk}
\textbf{Generative Adversarial Networks (GAN).}
Vanilla GAN \cite{goodfellow2014generative} consists of two key components, a generator ${\mathcal G}$, and a discriminator ${\mathcal D}$. Given a prior distribution $Z$ as input,  ${\mathcal G}$ maps an instance $\mathbf{z} \sim Z$ from the latent space to the data space as ${\mathcal G}(\mathbf{z})$.  On the other hand, ${\mathcal D}$ attempts to distinguish a data instance $\mathbf{x}$ from a synthetic instance ${\mathcal G}(\mathbf{z})$, generated by ${\mathcal G}$. We use the terminology $p_{Z}(\mathbf{z})$ to denote that $\mathbf{z}$ is a sampled instance from the distribution $Z$. The training process is set up as if ${\mathcal G}$ and ${\mathcal D}$ are playing a zero-sum game, a.k.a. minimax game; ${\mathcal G}$ tries to generate the synthetic instances that are as close as possible to actual data instances; on the other hand, ${\mathcal D}$ is responsible for distinguishing the real instances from the synthetic instances. In the end, GAN converges when both ${\mathcal G}$ and ${\mathcal D}$ reach a Nash equilibrium; at that stage, ${\mathcal G}$ learns the data distribution and is able to generate data instances that are very close to the actual data instances. The objective function of GAN can be written as follows:
\begin{equation}
\begin{split} 
\min_{\mathcal{G}} \max_{\mathcal{D}} V(\mathcal{D}, \mathcal{G}) = \underset{\mathbf{x} \sim {X}} {\mathbb{E}} [\log \mathcal{D}(\mathbf{x})] + 
\underset{\mathbf{z} \sim {Z}} {\mathbb{E}} [\log (1-\mathcal{D}(\mathcal{G}(\mathbf{z})))]
\end{split}
\end{equation}
where $X$ is the distribution of $\mathbf{x}$ and $Z$ is the distribution from which ${\mathcal G}$ samples.

\textbf{Bidirectional Generative Adversarial Networks (BiGAN).}
Besides training a generator ${\mathcal G}$, BiGAN \cite{donahue2017adversarial} also trains an encoder ${\mathcal E}$, that maps real instances $\mathbf{x}$ into latent feature space ${\mathcal E}(\mathbf{x})$. Its discriminator ${\mathcal D}$ takes both $\mathbf{x}$ and $p_{Z}(\mathbf{z})$ as input in order to match the joint distribution $p_{\mathcal{G}}(\mathbf{x}, \mathbf{z})$ and 
$p_{\mathcal{E}}(\mathbf{x}, \mathbf{z})$. The objective function of BiGAN can be written as follows:
\begin{equation}
\begin{split} 
\min_{\mathcal{G}, \mathcal{E}} \max_{\mathcal{D}} V(\mathcal{D}, \mathcal{E}, \mathcal{G}) = \underset{\mathbf{x} \sim X} {\mathbb{E}} [\log \mathcal{D}(\mathbf{x}, \mathcal{E}(\mathbf{x}))]   + \underset{\mathbf{z} \sim Z} {\mathbb{E}} [\log (1-\mathcal{D}(\mathcal{G}(\mathbf{z}), \mathbf{z}))]
\end{split}
\end{equation}
The objective function achieves the global minimum if and only if the distribution of both generator and encoder matches., i.e., $p_{\mathcal{G}}(\mathbf{x}, \mathbf{z}) = p_{\mathcal{E}}(\mathbf{x}, \mathbf{z})$.

\section{Methodology}
\label{sec:method}
\begin{figure*}[h]
  \centering
  \includegraphics[width=\linewidth]{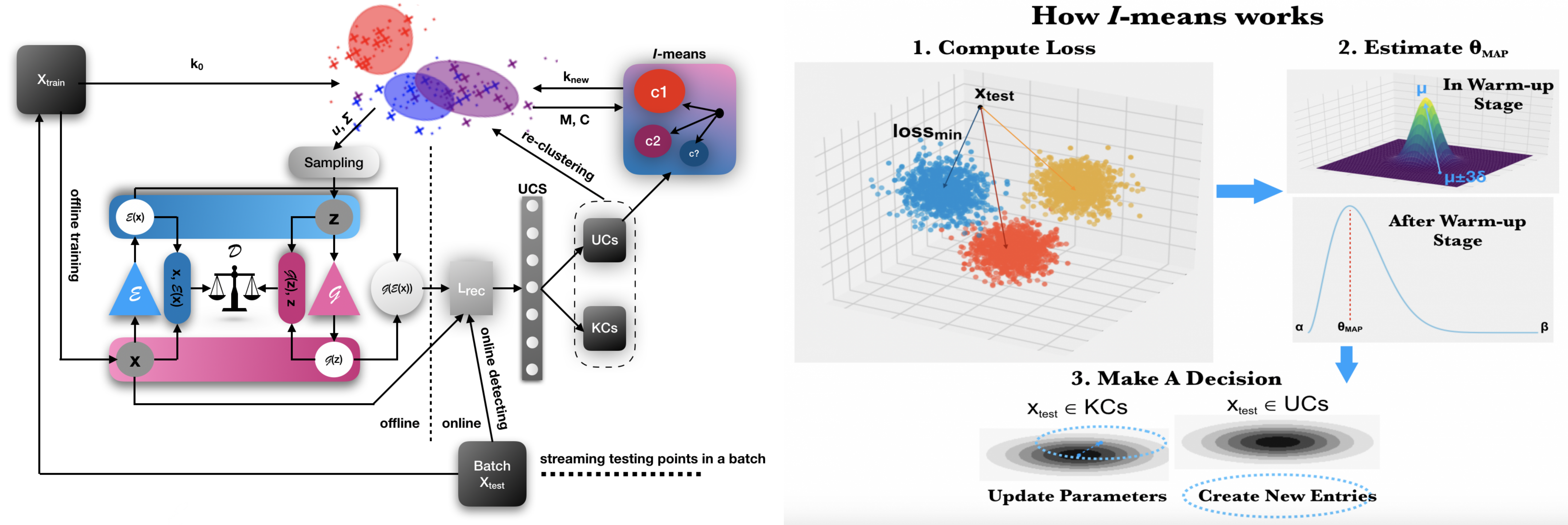}
  \caption{The Model Architecture of NE-GM-GAN (Left-hand Side) and The Workflow of $\textit{I}$-means in Algorithm (\ref{algo:I_means}) (Right-hand Side)}
\label{fig:model}
\end{figure*}

In this paper, we propose a novel model, Non-Exhaustive Gaussian Mixture Generative Adversarial Networks (NE-GM-GAN) for online non-exhaustive learning. The whole process is displayed in Figure \ref{fig:model}. Given a training set $X_{train}$ with $k_0$ KCs, in the training step (offline), the proposed NE-GM-GAN employs a bidirectional GAN to train its encoder ${\mathcal E}$ and generator ${\mathcal G}$, by matching the joint distribution of encoder $(X, Z)$ with the same of the generator. Note that the prior distribution $Z$ of ${\mathcal G}$ is a multi-modal Gaussian (shown as Gaussian clusters on the top-middle part of the figure). After training, the generator and encoder of the GAN can take $\mathbf{z}$ and $\mathbf{x}$ as input and generate ${\mathcal G}(\mathbf{z})$ and ${\mathcal E}(\mathbf{x})$ as output, respectively.

The test step (online) shown on the right side of the model architecture and it is run on a batch of input instances, $X_{test}$. For all data instance from a  batch (say, $\mathbf{x}$ is one such instance), NE-GM-GAN computes the $UCS(\mathbf{x})$ (unknown class score) of all instances in that batch; UCS score is derived from the reconstruction loss $L_{rec} = \lvert \mathbf{x} - \mathcal{G}(\mathcal{E}(\mathbf{x})) \rvert$. Using this score, the instances of a batch are partitioned into two groups: KCs and UCs. Elements in KCs belong to the known class, whereas the elements in UCs are potential UC instances. Using instances of UCs group, the model estimates the number of emerging class, $k_{new}$. After estimation, the model updates the prior of the ${\mathcal G}$ by adding the number of new classes $k_{new}$ to $k_0$ as shown in the top right part of the model architecture. The GMM is then retrained for clustering both KCs and UCs. At this stage, the online test process for one test batch is finished. 

In subsequent discussion, $X_{train} \in \mathbb{R}^{r \times d}$ is considered to be training data, containing $r$ data instances, each of which is represented as a $d$-dimensional vector. $K$ is the total number of known classes in $X_{train}$. $X_{test}$ is test data that may contain instances of KCs and also instances of UCs. The dimensionality of latent space is denoted by $p$.

\noindent
\textbf{Offline Training: Computing Multi-modal Prior Distribution} \\
In the vanilla form, generators of both GAN and BiGAN has a unimodal distribution as prior; in other words, the random variables $p_{Z}(\mathbf{z})$ is an instance from a unimodal distribution. Enlightened by \cite{benyosef2018gaussian}, in this paper, we consider a multi-modal distribution as prior since this prior can better fit the real-life distribution of multi-class datasets. Thus, 
\begin{equation}
p_{Z}(\mathbf{z}) = \sum_{k=1}^{K} \alpha^{\{k\}} \cdot p_{k}(\mathbf{z})
\end{equation}
We assume that the number of initial clusters in the Gaussian distribution matches with the number of known classes ($K$) in $X_{train}$. $\alpha^{\{k\}}$ is the mixing parameter, $p_{k}(\mathbf{z})$ denotes the multivariate Normal distribution $\mathcal{N}(u^{\{k\}}, \Sigma^{\{k\}})$, where $u^{\{k\}}$ and $\Sigma^{\{k\}}$ are mean vector and co-variance matrix, respectively.

The model assumes that the number of instances and the number of known classes in the training set are given at the beginning. During training (offline), the parameters $u^{\{k\}}$ and $\Sigma^{\{k\}}$ of each Gaussian cluster is learned by GMM and they are used as the sampling distribution of the latent variable for generating the adversarial instances. Suggested by \cite{benyosef2018gaussian}, we also use the re-parameterization trick in this paper. Instead of sampling the latent variable $\mathbf{z} \sim N(u^{\{k\}}, \Sigma^{\{k\}})$, the model samples $\mathbf{z} = A^{\{k\}}\epsilon + u^{\{k\}}$, where $\epsilon \sim N(0, I)$, $A \in \mathbb{R}^{p \times p}$, $u^{\{k\}} \in \mathbb{R}^{p}$. In this scenario, $u(\mathbf{z}) = u^{\{k\}}$ and $\Sigma(\mathbf{z}) =  A^{\{k\}} A^{\{k\}T}$.

Similar to \cite{donahue2017adversarial}, the GM-GAN (Gaussian Mixture-GAN) learning proceeds as follows. The model takes sampled instance $\mathbf{z}$, sampled from the Gaussian multi-modal prior and a real instances $\mathbf{x}$ as input. Generator ${\mathcal G}$ attempts to map this sampled  $p_{Z}(\mathbf{z})$  to data space as ${\mathcal G}(\mathbf{z})$. Encoder ${\mathcal E}$ maps real instances $\mathbf{x}$ into latent feature space as ${\mathcal E}(\mathbf{x})$. Discriminator ${\mathcal D}$ takes both $p_{Z}(\mathbf{z})$ and $\mathbf{x}$ as input for matching their joint distributions. After the model converges, theoretically, ${\mathcal G}(\mathbf{z}) \sim \mathbf{x}$ and ${\mathcal E}(\mathbf{x}) \sim p_{Z}(\mathbf{z})$.
Note that NE-GM-GAN encodes $X_{train}$ for offline training. To do so, GMM takes encoded $X_{train}$ as input and then generates encoded $u$ and $\Sigma$.

~\\
\noindent
\textbf{Extracting Potential Unknown Class} \\
UC extraction of NE-GM-GAN is an online process that works on unlabeled data. During online detection, the model assumes that the test instance $\mathbf{x}$ is coming in a batch of the test set $X_{test} \in \mathbb{R}^{b \times d}$, where $b$ is batch size and $d$ is the dimension of feature space. Unlike \cite{donahue2017adversarial}, whose purpose is to generate the fake images as real as possible, our model aims at extracting the UC as accurately as possible. More specifically, our model generates the reconstructed instance ${\mathcal G}({\mathcal E}(\mathbf{x}))$ at first and then computes the reconstruction loss between $\mathbf{x}$ and ${\mathcal G}({\mathcal E}(\mathbf{x}))$. This step returns a size-$b$ 1-D vector, consisting of reconstruction losses of the $b$ points in the current batch, 
which is defined below:
\begin{equation}
L_{rec} = \Vert \mathbf{x} - \mathcal{G}(\mathcal{E}(\mathbf{x})) \Vert
\label{eqn:lrec}
\end{equation}

To distinguish the UC from KC in each test batch, we propose a metric, unknown class score, in short, $UCS$; the larger the score for an instance, the more likely that the instance belongs to an unknown class. To compute $UCS$ of a test instance $\mathbf{x}$, NE-GM-GAN first computes, for each KC (out of $K$ KCs), a baseline reconstruction loss, which is equal to the median of reconstruction losses of all train objects belonging to that known class. Then, $UCS$ of $\mathbf{x}$ is equal to the minimum of the differences between $\mathbf{x}$'s reconstruction loss and each of the $K$ baseline reconstruction losses. The pseudo-code of $UCS$ computation is shown in Algorithm \ref{algo:ucs}.

The intuition of $UCS$ function is that GAN models instances of KCs with smaller reconstruction loss than the instances of UCs, but different known classes may have different baseline reconstruction loss, so we want an unknown class's reconstruction loss larger than the worst loss among all the KCs. This mechanism is inspired by \cite{zenati2018adversarially}. Nevertheless, unlike \cite{zenati2018adversarially}, which assumes the prior as unimodal distribution and the UC must be far away from KC, our approach considers a multi-modal prior. After computing the $UCS$, the model extracts the potential UC from KC with a given threshold. For online detection, the threshold for the first test batch is empirically given whereas subsequent thresholds are decided by the percentage of UCs from previous test batches. Note that, the UCs objects may belong to multiple classes, but the model has no knowledge yet about the number of classes.

\begin{algorithm}
\DontPrintSemicolon 
\KwIn{Matrix $X_{train} \in \mathbb{R}^{r \times d}$ and $X_{test} \in \mathbb{R}^{b \times d}$}
Compute $L_{test}(x_{test})$ with Equation (\ref{eqn:lrec}); \\
\For{$i \gets 1$  $\textbf{to}$  $b$} {
  \For{$k \gets 1$  $\textbf{to}$  $K$} {
    Compute $L_{train}(x_{train})^{\{k\}}$ with Equation (\ref{eqn:lrec}); \\
    Select the median of $L_{train}(x_{train})^{\{k\}}$; \\
    $UCS(x_{test})^{\{k\}}$ = $\left| L_{test}(x_{test})^{\{i\}} - L_{train}(x_{train})^{\{k\}}_{median} \right|$;
  }
  $UCS^{\{i\}}_{min} = \min \left(UCS(x_{test})^{\{1\}}, ..., UCS(x_{test})^{\{K\}} \right)$;
}
$UCS = [UCS^{\{1\}}_{min}, ..., UCS^{\{b\}}_{min}]$; \\
\Return{Vector $UCS \in \mathbb{R}^{b \times 1}$}
\caption{{\sc $UCS$} for multi-modal prior}
\label{algo:ucs}
\end{algorithm}

\noindent
\textbf{Estimating The Number of Emerging Class} \\
The previous extraction only extracts potential UCs. In practice, a small number of anomalous KC instances may be selected as UC instances. So, we use a subsequent step that distinctly identifies instances of unknown classes together with the number of UC and their parameters (mean, and covariance matrix of each of the UCs). We name this step as Infinite Means ($\textit{I}$-means); the name reflects the fact that the number of unknown classes can increase as large as needed based on the test instances. Using $\textit{I}$-means, a test instance is assigned to a new class if it is positioned far from the mean of all the KCs, and discovered novel classes prior to seeing that instance. To achieve this, for $i$-th test instance $x_{test}^{\{i\}}$, as shown in Equation (\ref{eqn:L_mu}), $\textit{I}$-means computes the distance $L_{\mu}^{\{k\}}$ between $x_{test}^{\{i\}}$ and the mean vector $\mu^{\{k\}}$ for the $k$-th KC and then selects the minimum of these values as $loss_{min}$ in Equation (\ref{eqn:L_min}).

{\belowdisplayshortskip=-3pt
\begin{equation}
L_{\mu}^{\{k\}} = \Vert x_{test}^{\{i\}} - \mu^{\{k\}} \Vert, \forall k \in [1..K]
\label{eqn:L_mu}
\end{equation}

\begin{equation}
\small
loss_{min} = \min \left(L_{\mu}^{\{1\}}, L_{\mu}^{\{2\}}, ..., L_{\mu}^{\{K\}}\right), \ 
idx = \arg\min \left(L_{\mu}^{\{1\}}, L_{\mu}^{\{2\}}, ..., L_{\mu}^{\{K\}}\right)
\label{eqn:L_min}
\end{equation}
}
A small value of $loss_{min}$ indicates that $x_{test}^{\{i\}}$ may potentially be a member of class $idx$; on the other hand, a large value $loss_{min}$ indicates that $x_{test}^{\{i\}}$ possibly belongs to a UC. To make the determination, we use a Bayesian approach, which dynamically adjusts the probability that a test point that is closest to cluster $idx$'s mean vector belongs to cluster $idx$ or not.
The process is described below.

For a test instance, $x_{test}^{\{i\}}$ for which $idx=k$, the binary decision whether the instance belongs to $k$-th existing cluster or an emerging cluster follows Bernoulli distribution with parameter $\theta_k$, which is modeled by using a Beta prior with parameter $\alpha_k$, and $\beta_k$, where $\alpha_k, \beta_k \ge 1$ and $\theta_k = \frac{\alpha_k}{\alpha_k+\beta_k}$. The value of $\alpha_k$ and $\beta_k$ are updated using Bayes rule. Based on the Bayes' theorem, the posterior distribution $p(\theta_k|x_{test}^{\{i\}})$, where $\theta_k \in [0, 1]$, is proportional to the prior distribution $p(\theta_k)$ multiplied by the likelihood function $p(x_{test}^{\{i\}}|\theta_k)$:
\begin{equation}
    p(\theta_k|x_{test}^{\{i\}}) \propto p(x_{test}^{\{i\}}|\theta_k) \cdot p(\theta_k)
\label{eq:bayes}
\end{equation}
The posterior $p(\theta_k|x_{test}^{\{i\}})$ in Equation (\ref{eq:bayes}) can be re-written as following:
\begin{equation}
\begin{aligned}
p(\theta_k|x_{test}^{\{i\}})
&\propto \theta_k^{\alpha_{k0}} (1-\theta_k)^{\beta_{k0}} \cdot \theta_k^{\alpha_k - 1} (1-\theta_k)^{\beta_k - 1} \\
&= \theta_k^{\alpha_{k0} + \alpha_k - 1} \cdot (1-\theta_k)^{\beta_{k0} + \beta_k - 1} \\
&= beta(\theta_k | \alpha_{k0} + \alpha_k, \beta_{k0} + \beta_k)
\end{aligned}
\end{equation}
As the test instances are coming in streaming fashion, for any subsequent test instance for which $idx=k$, the posterior $p(\theta_k|x_{test}^{\{i\}})$ will act as prior for the next update. For the very first iteration, $\alpha_{k0}$ and $\beta_{k0}$ are shape parameters of beta prior, which we learn in a warm-up stage. In the warm-up stage, we apply the three-sigma rule to compute the beta priors, $\alpha_{k0}$, and $\beta_{k0}$. Each test point in the warm-up stage, for which $idx=k$, contributes a count of 1 to $\alpha_{k0}$ if the point is further than $3$ standard deviation away from the mean, otherwise it contributes a count of 1 to $\beta_{k0}$.
After the warm-up stage, we employ the Maximum-A-Posteriori (MAP) estimation to obtain the $\theta_{MAP_k}$ at which the posterior $p(\theta_k|x_{test}^{\{i\}})$ reaches its maximum value. According to the property of beta distribution, the $\theta_{MAP_k}$ is most likely to occur at the mean of posterior $p(\theta_k|x_{test}^{\{i\}})$. Thus, we can estimate the $\theta_{MAP_k}$ by:
\begin{equation}
\begin{aligned}
\theta_{MAP_k} = \underset{\theta_k}{\arg \max} \ p(\theta_k | x_{test}^{\{i\}}) = \frac{\alpha_{k0} + \alpha_k}{\alpha_{k0} + \alpha_k + \beta_{k0} + \beta_k}
\end{aligned}
\label{eq:map}
\end{equation}

After estimating the $\theta_{MAP_k}$ by Equation (\ref{eq:map}), $\textit{I}$-means makes a cluster membership decision for each $x_{test}^{\{i\}}$ based on $\theta_{MAP_k}$. This decision simulates the Bernoulli process, i.e., among the test instances which are close to the $k$-th cluster, approximately $\theta_{MAP_k}$ fraction of those will belong to the emerging cluster, whereas the remaining ($1 - \theta_{MAP_k}$) fractions of such instances will belongs to the $k$-th cluster. After each decision, corresponding parameters will be updated. If $x_{test}^{\{i\}}$ is clustered as a member of $KC^{\{k\}}$, we update the parameters  $\mu_{k}^{\{i\}} \in \mathbb{R}^{1 \times d}$, $\sigma_{k}^{\{i\}} \in \mathbb{R}^{d \times d}$ of the $k$-th cluster by Equation (\ref{eq:update_mu}) and Equation (\ref{eq:update_sigma}), respectively. The shape parameter $\beta_k$ is increased by 1. Otherwise, if $x_{test}^{\{i\}}$ is considered as a member of UC, the shape parameter $\alpha_k$, $k_{new}$ are increased by 1, and the mean and covariance matrix of this new class are initialized by assigning current $x_{test}^{\{i\}}$ as new mean vector and creating a zero vector with the same shape of $x_{test}^{\{i\}}$ as new standard deviation vector.

{\belowdisplayshortskip=-6pt
\begin{equation}
    \mu_{k}^{\{i\}} = \mu_{k}^{\{i-1\}} + \frac{ x_{test}^{\{i\}} - \mu_{k}^{\{i-1\}} }{ i }
\label{eq:update_mu}
\end{equation}

\begin{equation}
v_{k}^{\{i\}} = v_{k}^{\{i-1\}} + \left( x_{test}^{\{i\}} - \mu_{k}^{\{i-1\}} \right) \left( x_{test}^{\{i\}} - \mu_{k}^{\{i\}} \right), \ 
\sigma_{k}^{\{i\}} = \sqrt{ \frac{v_{k}^{\{i\}}}{(i-1)} }
\label{eq:update_sigma}
\end{equation}
The entire process of this paragraph is summarized $\textit{I}$-means in Algorithm \ref{algo:I_means}.
}

\begin{table*}
\footnotesize
\centering
\setlength{\tabcolsep}{2pt}
  \caption{Statistics of Datasets (\#Inst. denotes the number of instances; \#F. denotes to the number of features after one-hot embedding or dropping for network intrusion dataset; \#C. denotes to the number of Classes.)}
  \label{tab:dataset}
  \begin{tabular}{ccccp{6.8cm}}
    \toprule
    \textbf{Dataset} & \textbf{\#Inst.} & \textbf{\#F.} & \textbf{\#C.} & \textbf{Selected UCs} \\
    \midrule
    \textbf{KDD99} & 494,021 & 121 & 23 & neptune, normal, back, satan, ipsweep, portsweep, warezclient, teardrop \\
    \textbf{NSL-KDD} & 148,517 & 121 & 40 & neptune, satan, ipsweep, smurf, portsweep, nmap, back, guess\_passwd \\
    \textbf{UNSW-NB15} & 175,341 & 169 & 10 & generic, exploits, fuzzers, DoS, reconnaissance, analysis, backdoor, shellcode \\
    \textbf{Synthetic} & 100,300 & 121 & 16 & No.3, 4, 5, 6, 7, 8, 9, 10 \\
  \bottomrule
\end{tabular}
\end{table*}

\begin{algorithm}
\DontPrintSemicolon 
\KwIn{Testing batch $X_{test} \in \mathbb{R}^{b \times d}$,
mean matrix,~co-variance matrix
}
\For{$\textbf{all} \ x^{\{i\}} \in X_{test}$} {
  \For{$\textbf{all} \ \mu^{\{k\}} \in M$} {
    Compute $L_{\mu}^{\{k\}}$ by Equation ($\ref{eqn:L_mu}$);
  }
  Get the index, $idx$, of minimum loss by Equation ($\ref{eqn:L_min}$); \\
  \If {\bf warm-up stage}{
    Select beta prior $\alpha_{idx0}$ and $\beta_{idx0}$ based on Three-sigma Rule;
  }
  \Else{
    Estimate the $\theta_{MAP_k}$ by Equation (\ref{eq:map});
  }
  \If {\bf Uniform $(0, 1) \le \theta_{MAP_k}$}{
    Update corresponding $\mu$ and $\sigma$ by Equation (\ref{eq:update_mu}) and Equation (\ref{eq:update_sigma}); \\
    $\beta_{idx} \gets \beta_{idx} +1;$
  }
  \Else{
    $\alpha_{idx} \gets \alpha_{idx} +1;$ \\
    $k_{new} \gets k_{new} +1;$
  }
}
\Return{The number of new emerging clusters $k_{new}$}
\caption{{\sc Infinite} Means ($\textit{I}$-means)}
\label{algo:I_means}
\end{algorithm}

\section{Experiments}
\label{sec:exp}
In this section, we show experimental results for validating the superior performance of our proposed NE-GM-GAN over different competing methods for multiple capabilities. Firstly, we compare the performance of potential UCs extraction. Furthermore, we compare the estimation of the number of distinct unknown classes. Finally, we show some experimental results for studying the effect of user-defined parameters on the algorithm's performance.

\noindent
\textbf{Dataset.}
We evaluate NE-GM-GAN on four datasets. Three of the datasets are real-life network intrusion datasets and the remaining one is a synthetic dataset. The network intrusion is very common for non-exhaustive classification because attackers constantly update their attack methods, so the classification model must adapt to novel class scenarios. The datasets are:
(1) KDD Cup 1999 network intrusion dataset \textbf{(KDD99)}, which contains 494,021 instances and 41 features with 23 different classes. One of the class represents “Normal” activity and the rest 22 represent various network attacks; (2) NSL-KDD dataset \textbf{(NSL-KDD)} \cite{dhanabal2015study}, which is also a network intrusion dataset built by filtering some records from KDD99; 
(3) UNSW-NB15 dataset \textbf{(UNSW-NB15)} \cite{moustafa2015unsw}, which hybridizes real normal network activities with synthetic attack; 
(4) Synthetic dataset \textbf{(Synthetic)}, which contains non-isotropic Gaussian clusters. 
Many of the features in the intrusion datasets are categorical or binary, so we employ one-hot embedding for such features. We also drop some columns which are redundant or whose values are almost zero or missing along the column. After that, we select eight of the classes as unknown classes (UCs) for each dataset.
The test set is constructed from two parts. The first part is randomly sampled 20\% of KCs instances and the second part is all the instances of the UCs. Rest 80\% of KC instances are left for the training set.
In the synthetic dataset, noises are injected into Gaussian clusters, each cluster representing a class. The injected noise is homocentric to the corresponding normal class but with a larger variance. The detailed statistics of the datasets are provided in Table \ref{tab:dataset}.

\noindent
\textbf{Competing Methods.}
The performance of UCs extraction is evaluated with three competing methods, AnoGAN \cite{schlegl2017unsupervised}, DAGMM \cite{zong2018deep}, and ALAD \cite{zenati2018adversarially}. AnoGAN is the first GAN-based model for UC detection. Similarly, ALAD is another GAN-based model, which uses reconstructed errors to determine the UC. In contrast, DAGMM implements the autoencoder for the same task instead. The experimental setting follows \cite{zenati2018adversarially} for this experiment.
On the other hand, the capability of estimating the number of new emerging classes is compared against two competing methods, X-means \cite{pelleg2000x}, and IGMM \cite{rasmussen2000infinite, zhang2016bayesian}. X-means is a classical distance-based algorithm that can efficiently search the data space without knowing the initial number of clusters. On the contrary, IGMM is a Bayesian mixture model which uses the Dirichlet process prior and Gibbs sampler to efficiently identify new emerging entities. This experiment uses one sweep Gibbs sampler for IGMM \cite{zhang2016bayesian}. For IGMM, we select the tunable parameters as following; $h=10$, $m=h+100$, $\kappa=100$ and $\alpha=100$, which is identical to the parameter values in \cite{zhang2016bayesian}. Both models can return the number of online classes as NE-GM-GAN does, so they are selected as competing methods.

\noindent
\textbf{Evaluation Metrics.}
We use an external clustering evaluation metric, such as F1-score, to evaluate the performance of UCs extraction. For evaluating the prediction of the number of UCs (a regression task), we propose a new metric, Symmetrical R-squared ($S$-$R^{2}$). To obtain this, the root mean square error ($RMSE$) for both NE-GM-GAN and a competing method are computed and plugged into Equation \ref{eqn:sr2}. $S$-$R^{2}$ $\in [\textnormal{-}1, 1]$ gets more close to 1 if NE-GM-GAN defeats the competing method. On the contrary, its value will become more close -1. $S$-$R^{2}$ is exactly equal to 1 when the proposed model gets perfect prediction while the competing method doesn't. $S$-$R^{2}$ is zero when both methods have similar performance. The motivation to propose a new metric rather than using R-squared ($R^{2}$) is that $R^{2}$ would be less distinctive if two methods get much worse predictions because of using mean square error ($MSE$) inside. Besides, baseline sometimes achieves better performance, but $R^{2}$ cannot reflect this scenario as its range is from negative infinity to positive one.

\begin{equation}
S \textnormal{-} R^{2}=\left\{
\begin{aligned}
1 - \frac{RMSE_{m}}{RMSE_{bl}} & , & RMSE_{m} < RMSE_{bl} \\
\frac{RMSE_{bl}}{RMSE_{m}} - 1 & , & RMSE_{m} > RMSE_{bl}
\label{eqn:sr2}
\end{aligned}
\right.
\end{equation}
where $RMSE_{m}$ and $RMSE_{bl}$ denote the $RMSE$ of our model and baseline model, respectively.

\noindent
\textbf{The Capability of Unknown Class Extraction.}
In Table \ref{table:extraction}, we show the F1-score values of NE-GM-GAN and the competing methods for detecting the unknown class instances (the best results are shown in bold font). The result is computed by running each model 10 times and then taking the average. Out of the four datasets, NE-GM-GAN has the best performance in three with a healthy margin over the second-best method. In the largest dataset, our model received a 0.99 F1-score, a very good performance considering the fact that unknown class instances are assembled from 8 different classes. Only in the NSL-KDD dataset, NE-GM-GAN came out as the second-best. The performance of the other three models is mixed without a clear winner. One observation is that all the methods perform better on the larger dataset (KDD99).

To understand NE-GM-GAN's performance further, we perform an ablation study by switching the prior, as shown in Table \ref{tab:prior}. As we can see Gaussian multi-modal prior used in NE-GM-GAN is better suited than Unimodal prior generally used in traditional GAN. For all datasets multi-modal prior has 1\% to 2\% better F-score. A possible reason is that multi-modal prior is more closer to the real distribution of the training data.

{
\setlength{\parskip}{2.0ex}

\begin{table}
\centering
\setlength{\tabcolsep}{3pt}
\caption{The F1-score of Four Models for UCs Extraction}
\label{table:extraction}
\begin{tabular}{c|c|c|c|c} 
\hline
\textbf{Data} & \textbf{NE-GM-GAN} & \textbf{AnoGAN} & \textbf{DAGMM} & \textbf{ALAD} \\
\hline
\textbf{KDD99} & \textbf{0.99} & 0.87 & 0.97 & 0.94 \\
\hline
\textbf{NSL-KDD} & 0.75 & 0.68 & \textbf{0.79} & 0.73 \\
\hline
\textbf{UNSW-NB15} & \textbf{0.57} & 0.49 & 0.53 & 0.51 \\
\hline
\textbf{Synthetic} & \textbf{0.74} & 0.51 & 0.70 & 0.56 \\
\hline
    \end{tabular}
\end{table}

\begin{table}
\centering
\setlength{\tabcolsep}{3pt}
  \caption{F1 Score from Our Proposed Model by Using Different Prior}
  \label{tab:prior}
  \begin{tabular}{ccccc}
    \toprule
    \textbf{Prior} & \textbf{KDD99} & \textbf{NSL-KDD} & \textbf{UNSW-NB15} & \textbf{Synthetic} \\
    \midrule
    \textbf{Unimodal} & 0.98 & 0.74 & 0.55 & 0.72 \\
    \textbf{Multi-modal} & \textbf{0.99} & \textbf{0.75} & \textbf{0.57} & \textbf{0.74} \\
  \bottomrule
\end{tabular}
\end{table}
}

\noindent
\textbf{The Estimation of The Number of New Classes.}
In this experiment, we compare NE-GM-GAN against two competing methods on all four datasets. To extend the scope of experiments, we vary the number of unknown classes from 2 to 6 by choosing all possible combinations of UCs and build multiple copies of one dataset and report performance results over all those copies. The motivation of using a combination of different UCs is to validate the robustness of the methods against varying numbers of UC counts.
The result is shown in Table \ref{table:performance} using $S$-$R^2$ metric discussed earlier. The result close to $1$ (the majority of the values in the table are between 0.85 and 0.95) means NE-GM-GAN substantially outperforms the competing methods. 
We argue that both competing methods assume that data distribution in each class follows mixture of Gaussian and thus fail to achieve good performance on realistic datasets. In only one dataset (Synthetic), X-means was able to obtain identical performance as ours' method, as both methods have the perfect prediction.

The same results are also shown in Figure \ref{fig:fig_bar} as bar charts. In this Figure, $y$-axis is the number of predicted clusters, and each group of bars denotes the number of actual clusters for different methods. As we can see, NE-GM-GAN's prediction is very close to the actual prediction, whereas the results of the completing methods are way-off, except for the X-means method on the synthetic dataset. These experimental results demonstrate that our NE-GM-GAN outperforms the competing methods in terms of accuracy and robustness.

{\setlength{\parskip}{3.0ex}
\begin{table*}
\small
\centering
\setlength{\tabcolsep}{5pt}
\caption{The $S$-$R^{2}$ between NE-GM-GAN and Baselines on 4 Datasets (We denote ``UCs'' as the number of unknown classes in this table)}
\label{table:performance}
\begin{tabular}{c|c|c|c|c|c|c} 
\hline
\textbf{Datasets} & \textbf{Methods} & \textbf{UCs=2} & \textbf{UCs=3} & \textbf{UCs=4} & \textbf{UCs=5} & \textbf{UCs=6} \\
\hline
\multirow{2}{*}{\textbf{KDD99}} & X-means & 
0.8301 & 0.8805 & 0.8628 & 0.9105 & 0.8812 \\
\cline{2-7} & IGMM & 
0.9528 & 0.8991 & 0.8908 & 0.9303 & 0.9248 \\
\hline
\multirow{2}{*}{\textbf{NSL-KDD}} & X-means & 
0.8892 & 0.8604 & 0.9539 & 0.9228 & 0.9184 \\
\cline{2-7} & IGMM & 
0.8771 & 0.8647 & 0.9517 & 0.9285 & 0.9238 \\
\hline
\multirow{2}{*}{\textbf{UNSW-NB15}} & X-means & 
0.8892 & 0.8604 & 0.9539 & 0.9228 & 0.9184 \\
\cline{2-7} & IGMM & 
0.8771 & 0.8647 & 0.9517 & 0.9285 & 0.9238 \\
\hline
\multirow{2}{*}{\textbf{Synthetic}} & X-means & 
0 & 0 & 0 & 0 & 0 \\
\cline{2-7} & IGMM & 
1 & 1 & 1 & 1 & 1 \\
\hline
    \end{tabular}
\end{table*}

\begin{figure}[h]
  \centering
  \includegraphics[width=\linewidth]{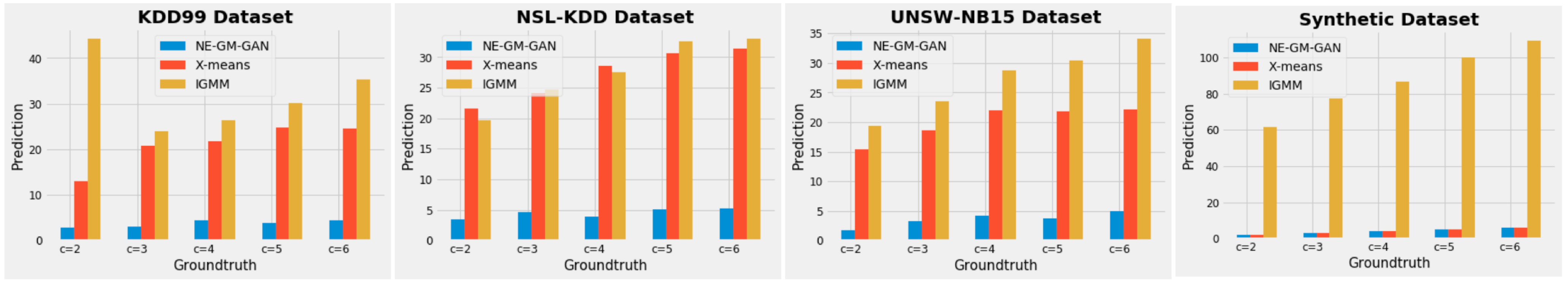}
  \caption{Comparison on The Estimation of New Emerging Class among Three Methods}
\label{fig:fig_bar}
\end{figure}
}

\noindent
\textbf{Study of User-defined Parameters.}
We perform a few experiments to justify some of our parameter design choices. For instance, to build the initial beta priors we used three-sigma rule. In Table \ref{tab:3sigma}, we present the percentage of instances of points that falls within the three standard deviations of the mean. The four columns correspond to the four datasets. As can be seen in the third row of the table, for all datasets, almost 100\% of the points falls within the three standard deviations away from the mean. So, the priors selected in the warm-up stage based on three-sigma rule can sufficiently distinguish the UCs from the known class instances.

\begin{figure}[h]
  \centering
  \includegraphics[width=\linewidth]{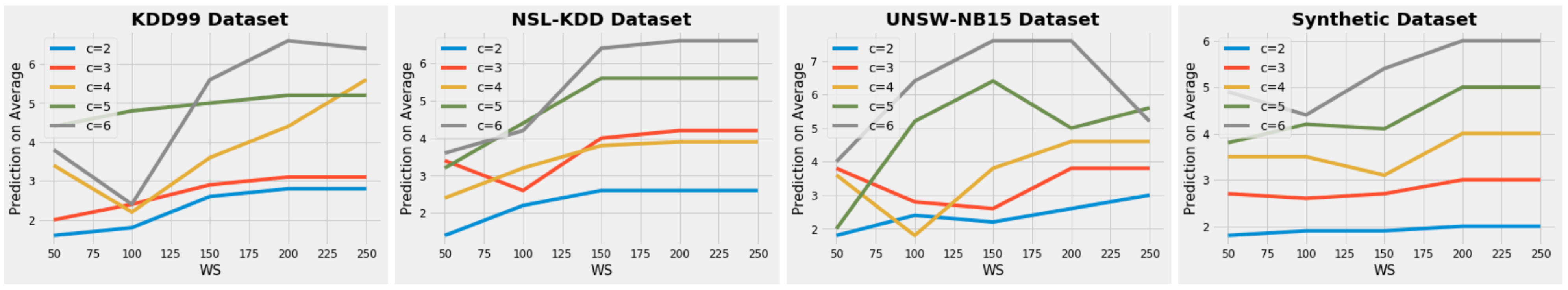}
  \caption{Investigation on The Number of Epochs in The Warm-up Stage ($WS$) for $\textit{I}$-means on Four Datasets}
\label{fig:fig_line}
\end{figure}

We also show unknown class prediction results over different values of $WS$ (epochs of the warm-up stage) for different (between 2 to 6) unknown class that counts for all datasets. In Figure \ref{fig:fig_line}, each curve represent a specific UC count. As can be seen, the prediction of the unknown class gets better with a larger number of $WS$. In most cases, the prediction converges when the number of epochs in the warm-up stage ($WS$) reaches 200 or above. In all our experiments, we select the $WS$ value 200 for all datasets.

\begin{table}
\centering
\setlength{\tabcolsep}{4pt}
  \caption{Test of Three-sigma Rule (\%)}
  \label{tab:3sigma}
  \begin{tabular}{ccccc}
    \toprule
    \textbf{Range} & \textbf{KDD99} & \textbf{NSL-KDD} & \textbf{UNSW-NB15} & \textbf{Synthetic} \\
    \midrule
    \textbf{$\mu \pm 1\sigma$} & 94.37 & 61.17 & 58.67 & 56.64 \\
    \textbf{$\mu \pm 2\sigma$} & 99.58 & 99.87 & 99.92 & 99.81 \\
    \textbf{$\mu \pm 3\sigma$} & 99.60 & 100.00 & 100.00 & 100.00 \\
  \bottomrule
\end{tabular}
\end{table}

\noindent
\textbf{Reproducibility of the Work.}
The model is implemented using Python 3.6.9 and Keras 2.2.4. For optimization, Adam is used with $\alpha=10^{-5}$ and $\beta= 0.5$; mini-batch size is 50, latent dimension is 32, and the number of training epochs equal to 1000. The source code is available at  \url{https://github.com/junzhuang-code/NEGMGAN}. The details of the BiGAN model architecture is given in Table~\ref{tab:model}.

\begin{table}[!t]
\footnotesize
\centering
\setlength{\tabcolsep}{6pt}
  \caption{Model Architectures}
  \label{tab:model}
  \begin{tabular}{cccccc}
    \toprule
     & \textbf{Layers} & \textbf{Units} & \textbf{Activation} & \textbf{Batch Norm.} & \textbf{Dropout} \\
    \midrule
    ${\mathcal E}(\mathbf{x})$ & Dense & 64 & LReLU(0.2) & $\times$ & 0.0 \\
    & Dense & 1 & None & $\times$ & 0.0 \\
    ${\mathcal G}(\mathbf{z})$ & Dense & 64 & LReLU(0.2) & $\times$ & 0.0 \\
    & Dense & 128 & LReLU(0.2) & $\times$ & 0.0 \\
    & Dense & 121 & Tanh & $\times$ & 0.0 \\
    ${\mathcal D}\mathbf{(x, z)}$ & Dense & 128 & LReLU(0.2) & $\checkmark$ & 0.5 \\
    & Dense & 128 & LReLU(0.2) & $\checkmark$ & 0.5 \\
    & Dense & 1 & Sigmoid & $\times$ & 0.0 \\
  \bottomrule
\end{tabular}
\end{table}

\section{Acknowledgement} This research is partially supported by National Science Foundation with grant number IIS-1909916.

\section{Conclusion}
\label{sec:con}
In this paper, we propose a new online non-exhaustive model, Non-Exhaustive Gaussian Mixture Generative Adversarial Network (NE-GM-GAN), that synthesizes Bayesian method and deep learning technique for incremental learning the new emerging classes. NE-GM-GAN consists of three main components: (1) Gaussian mixture clustering generating multi-modal prior and re-clusters both KCs and UCs for parameter updating. (2) Bidirectional adversarial learning reconstructs the loss for extracting imbalanced UCs from KCs in an online testing batch. (3) A novel algorithm, $\textit{I}$-means, estimates the number of new emerging classes for incremental learning the UCs on large sparse datasets. Experimental results illustrate that NE-GM-GAN significantly outperforms the competing methods for online detection across several benchmark datasets.

\bibliographystyle{splncs04}
\bibliography{reference}


\end{document}